# Machine Learning Based Mobile Network Throughput Classification

Lauri Alho
Computing Sciences,
Tampere University,
Tampere, Finland
lauri.alho@tuni.fi

Adrian Burian
Nokia Software,
Tampere, Finland
adrian.1.burian@nokia.com

Janne Helenius
Nokia Software,
Tampere, Finland
janne.helenius@nokia.com

Joni Pajarinen
Computing Sciences,
Tampere University,
Tampere, Finland
joni.pajarinen@tuni.fi

*Abstract*— **Identifying mobile network problems in 4G cells is more challenging when the complexity of the network increases, and privacy concerns limit the information content of the data. This paper proposes a data driven model for identifying 4G cells that have fundamental network throughput problems. The proposed model takes advantage of clustering and Deep Neural Networks (DNNs). Model parameters are learnt using a small number of expert-labeled data. To achieve case specific classification, we propose a model that contains a multiple clustering models block, for capturing features common for problematic cells. The captured features of this block are then used as an input to a DNN. Experiments show that the proposed model outperforms a simple classifier in identifying cells with network throughput problems. To the best of the authors' knowledge, there is no related research where network throughput classification is performed on the cell level with information gathered only from the service provider's side.**

## I. INTRODUCTION

We are living in an era where communication between radio devices is more important than ever before. The importance creates constantly increasing technical requirements for Mobile Networks (MNs) and User Equipment (UE), and progress can be seen in cases such as increased network reliability, lowered latency and increased network throughput [1]. The first mobile phone network generation, 1G, was released in 1979 [2]. After that, four major new generations have been released, 5G as the latest.

Technological improvements are required to create more diverse and complex networks [1]. These networks are usually built to achieve carrier class reliability, meaning 99.999 % network availability [3]. To achieve this, the increased number of UE, cells and base stations in MNs require new solutions for more efficient monitoring [2]. In mobile network monitoring, identifying problems while taking data privacy into account can be challenging due to the limited amount of informative data.

Machine Learning (ML) methods can be used to detect anomalies, e.g. network throughput problems, by using a dataset gathered on the cell level in the network. Understanding downlink throughput problems on the cell level is crucial for MN owners, because poorly behaving cells can decrease drastically users' Quality of Service (QoS) [2],[4].

Identifying the state of a cell requires usually human interaction. Classifying continuously a high number of cells to working and problematic ones can be unmanageable for a human. To overcome the challenges of the mobile network operators for identifying problematic cells, we propose an unsupervised and supervised learning method using a small number of expert-labelled data to identify throughput problematic cells in 4G networks. This is done by using clustering algorithms and Deep Neural Networks (DNNs). The proposed method classifies cells by their condition, which is either normal or problematic. The used dataset contains only monitoring data between a cell and served User Equipment (UE), and the true state of a cell identified by an expert. This means the expert only needs to label the cells during model training, and after training the model can classify all new cells on its own.

The main research question we try to answer is the following: Can throughput problematic cells be identified from network monitoring data with a low number of labels?

The main contributions in this paper are the following:

1. We propose a novel ML based approach that classifies throughput problematic cells using monitoring data with a low number of labels.

2. The proposed method has better performance than a simple classifier baseline.

3. Related research has used data gathered from UE in order to provide more accurate predictions for the throughput. This paper proposes a method which uses only network monitoring data from the MN side.

The rest of the paper is organized as follows. Section II presents the related work in the MN domain. Section III describes the 4G mobile network environment whereof the dataset is gathered and methods that are used for data pre-processing and data augmentation. Section IV describes the proposed model architecture which is split into a clustering block and a DNN. Section V shows experiments and results with the proposed method and the baseline. Section VI concludes this paper.

## II. RELATED WORK

ML has been applied successfully in the area of mobile networks, including assisted mobile network planning [5], [6], techniques for analysing big data in 5G network [7], and anomaly detection in an operating network [8]. While some of the research focuses on planning MNs efficiently, the proposed method in this paper is closer to solutions which aim to optimize and identify problems in operating networks. Those studies focus on areas such as throughput prediction, e.g. [9], [10] and [11], and the identification of MN problems, e.g. [4], [12], and [13].

Throughput prediction was studied in [11] showing it was possible to predict the throughput of UE by combining mobile network and user level data. Our proposed method in this paper differs from [11] by using only network level features and therefore the need of non-MN specific information, e.g. GPS location, are not used. In addition, [11] focuses on only predicting the future downlink throughput of the UE, but there is no classification made for identifying throughput problems. This is the biggest difference between [11] and this paper. Other differences are that we combine unsupervised and supervised learning motivated by the very small number of labels in the dataset. We are not the aware of prior research in classifying cells based on the throughput.

Detecting and predicting MN outages with log data was studied in [12]. The main contribution was to develop a novel method for detecting and predicting anomalies in cellular networks nearly real time. The work focused on finding statistical differences between normal and outage periods in cellular networks. However, in our case, it is not sufficient to use only unsupervised learning since our dataset contains also labelled data by an expert. Moreover, the nature of the problem is different, as our problem is related to identifying problems in situations when the signal quality from a cell to UE is normal, but the throughput is low.

### III. ENVIRONMENT AND DATASET

#### A. 4G Mobile Network

Components of every MN generation can be divided into Radio Access Network (RAN) and Core Network. RAN components include functionality for air interface access between UE and Core Network, and Core Network provides centralized functions for e.g. user authentication and access to the Internet.

4G MNs use LTE architecture with main changes to previous generation mobile network architectures (1G, 2G and 3G) being the new Core Network component Evolved Packet Core (EPC). It contains five distinct modules, which are the Mobility Management Entity (MME), the Home Subscriber Server (HSS), Serving Gateway (S-GW), the Packet Data Network Gateway (P-GW) and the Policy Control and Charging Rules Function (PCRF) [1]. Fig. 1 illustrates the connections between the components of the LTE network architecture where orange arrows indicate control plane signals and blue arrows indicate user plane signals.

Control plane signals are used e.g. when a connection is setup between UE and a cell, or UE changes to another cell (handover). User plane signals are used to deliver the data packets between UE and the Internet.

When UE is connecting to an MN, an authentication request is sent to HSS by MME to authenticate the UE. If authentication is successful, UE initializes a user plane connection with the S-GW. S-GW is required to create a connection between EPC and RAN. The P-GW is responsible for connecting EPC to internet, and it uses PCRF to determine Quality-of-Service enforcement.

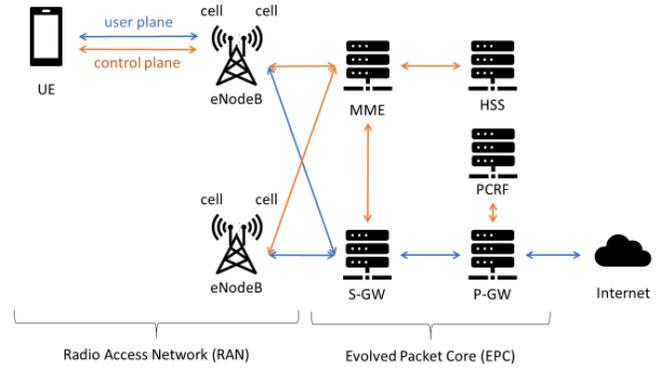

Fig. 1. Overview of LTE network architecture. Blue arrows indicate user plane signals and orange arrows control plane signals between components.

To enhance the data transmission speed between UE and a cell, carrier aggregation can be used in the air interface. When carrier aggregation is used, one primary cell and one or more secondary cells are used for data transferring [14]. In this work, only throughput of the primary serving cell of the UE is measured in the used dataset. The data is gathered from eNodeB including both the user plane and control plane traffic.

#### B. Dataset

The dataset for predicting throughput problems can be collected from a 4G mobile network. Every sample of the dataset contains information about the transmission between a single UE and its serving cell. Data features contain summarized and averaged information of the network state, such as a four second average of the downlink throughput of the UE and its serving cell. The most of the features in the dataset are standardized in LTE architecture. Table I describes the features in more detail including a short description, is it standardized, a possible unit, value range and example value.

The dataset we are using has 25 591 samples. It is divided into about 70 % for the training set and 30 % for the test set. The true state of every cell in the training and test set is identified by an expert and the corresponding label is given for every sample. In addition, another split with about 90 % of samples belonging for training set and 10 % into test set is created and labeled by the expert in the same way. The expert is only needed during model training to give correct states of the cells. The model is able to identify new problematic cells on its own after the model is trained.

#### C. Data Pre-Processing

The dataset is pre-processed before it is used for model training and testing. Pre-processing is done in the following order.

1. Cells which have lower than 100 samples in the training set are filtered out. This is done to ensure that enough real samples are available during model training.

TABLE I. SUMMARY OF THE DATASET FEATURES

| Feature | Standardized in LTE | Description | Unit | Value range | Example value |
|---|---|---|---|---|---|
| Cell Data Rate | no | Four second average of traffic throughput of the serving cell. | kilobit per second (kbps) | 0– | 1 200.0 kbps |
| Cell Identifier | no | Cell identification number. Specifies primary serving cell of the UE. | - | - | 2 |
| Channel Quality Indicator (CQI) | yes | Four second average of the channel quality from the serving cell to the UE | - | 0–15 | 6 |
| E-UTRAN Radio Access Bearer (ERAB) duration | yes | Time of the session has been alive between the UE and its serving cell. | second (s) | 0– | 10 s |
| Frequency Band | yes | The used LTE frequency band coded as a single number. | - | - | 1 |
| Load Active | no | Four second average of the number of active UE in the serving cell. | - | 0– | 15 |
| Physical Resource Block (PRB) | yes | Four second average of the number of Physical Resource Blocks used by the UE in Transmission Time Interval (TTI). Maximum PRB number depends on the used bandwidth which is in all cells smaller than 10 MHz. | - | 0–100 | 10 |
| Reference Signal Received Power (RSRP) | yes | The power of the reference signal from the cell which is measured by the UE. | decibel-milliwatt (dBm) | -44–140 | -20 dBm |
| Reference Signal Received Quality (RSRQ) | yes | The quality of the reference signal from the cell which is measured by the UE. | decibel (dB) | -19.5–(-3) | -8 dB |
| Speed Range | no | Average speed of the UE during the last minute. | - | 1, 2 or 3 | 1 |
| Throughput | no | Four second average of the UE downlink throughput of the primary serving cell. Throughput is measured from L2/MAC layer. | kilobit per second (kbps) | 0– | 1 000.5 kbps |
| Time Interval | no | Generation time of the sample in 6 hours accuracy. | - | 1, 2, 3 or 4 | 2 |
| Timing Advance (TA) | yes | Distance between UE and cell. Granularity is 78 meters. | meter (m) | 0– | 312 m |
| Transmission Time Interval (TTI) | yes | Four second average of the downlink PRB use per TTI. TTI is a 1 millisecond long period, where maximum number of PRBs is 100. | - | 0–100 | 20 |
| UE Identifier | no | Date-specific identification number of the UE. | - | - | 1 |

2. Equation (1) is used for converting RSRP values from logarithmic units into linear

$$RSRP_{linear} = 10\wedge(RSRP_{dBm} / 10) / 1\,000 \quad (1)$$

where $RSRP_{dBm}$ is the RSRP value in dBm unit.

3. RSRQ values are converted from logarithmic units to linear with (2) which is defined as

$$RSRQ_{linear} = 10\wedge(RSRQ_{dB} / 20) \quad (2)$$

where $RSRQ_{dB}$ is the RSRQ value in logarithmic format.

4. Normalization is done by using (3) which is min-max feature scaling. Equation (3) is defined as

$$F_{norm} = [F - \min(F)] / [\max(F) - \min(F)] \quad (3)$$

where $F$ is the one-dimensional vector, min is the function which returns the minimum value of the given argument, and max is the function which returns the maximum value of the given argument. Normalization is done for every feature except Cell Identifier and UE Identifier which are scrambled.

### D. Data Augmentation

The sample count of cells varies in the dataset. Because the input shape of the proposed model is fixed, data augmentation is used to generate a fixed number of samples for cells which have too few samples. The new samples are generated by duplicating the samples of the cell.

If a cell has more samples than the proposed model can use, the number of the samples in the cell is decreased by dropping randomly samples out.

## IV. MODEL ARCHITECTURE

The model architecture is divided into two sections, the clustering block and the DNN. Firstly, the clustering block containing multiple clustering models is created and trained with unsupervised learning. Secondly, the output of the clustering block is used as an input for the DNN, which is trained with supervised learning.

### A. Clustering Block

The clustering block contains multiple clustering models which are trained with unsupervised learning. Fig. 2 shows the overview of the clustering block.

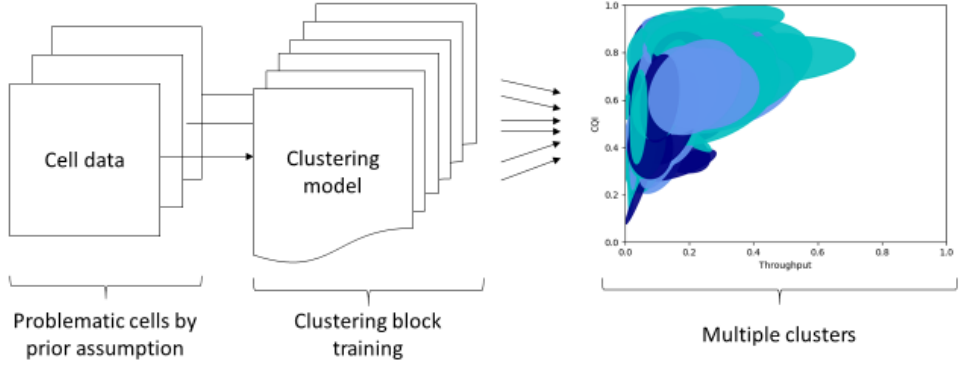

Fig. 2. Illustration of clustering block training with assumed problematic cells which selection is shown in Fig. 3. We train clustering models with assumed problematic cells as those probably contain the features shared with actual problematic cells. The clustering block is used to train the DNN, as illustrated in Fig. 4.

The number of created clustering models is defined by the number of problematic cells identified in the training set by the prior assumption. Selecting problematic cells with prior assumption is done as follows:

1. Calculate average of throughput ($T_{Avg,\ UE}$) and average of CQI ($C_{Avg,\ UE}$) by UE of each cell.
2. Calculate cell specific average of $T_{Avg,\ UE}$ values ($T_{Avg,\ Cell}$) and cell specific average of $C_{Avg,\ UE}$ values ($C_{Avg,\ Cell}$).
3. Create an intersection of the highest 30 % of $C_{Avg,\ Cell}$ values and the lowest 30 % of $T_{Avg,\ Cell}$ values. Cells in the intersection are used for training the clustering block.

Fig. 3 illustrates the intersection of the cells, which are identified by the algorithm. It is worth to note that this is only prior assumption, which makes the classification easier for the DNN.

Table II shows the used hyperparameters for the clustering block. We use K-Means and Gaussian Mixture Model algorithms to cluster samples into 2–10 clusters. This means that for every problematic cell identified by the prior assumption algorithm, we create and train 18 clustering models with different parameters.

All features except UE and cell identifier are used as an input for the clustering models. The weights of the clustering models are not changed after the training.

The output of a clustering model is a one-dimensional vector which is one-hot encoded. Every clustering model must be one-hot encoded independently, because the different parameters used for training the models changes the maximum cluster count and thus the output size of the one-hot encoding. The output of the clustering block is two-dimensional array, which is flattened and then used as an input for the DNN.

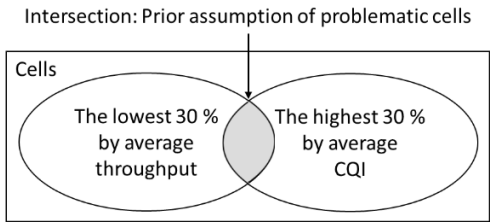

Fig. 3. Illustration of selecting the problematic cells based on the prior assumption in the training set.

### B. Deep Neural Network

The DNN is a binary classifier which uses flattened one-hot encoded output of the clustering block as an input. Fig. 4 presents the architecture of the model. The number of hidden layers and units per layer is defined by selected hyperparameters which are shown in Table II. Every hidden layer except the output layer has a ReLu activation function. The output layer has softmax activation function and additionally L1-regularization. The loss function is Mean Squared Error (MSE) and optimizer is Stochastic Gradient Descent (SDG) with 0.001 learning rate. If *decreasing units* hyperparameter is true, units of every layer is calculated with the following equation

$$U_l = U_{l-1} / 2^l \qquad (4)$$

where $l$ is the index of the layer, $U_l$ is the number of units of the $l$:th layer, and $U_1$ is defined by the hyperparameter *units per layer*.

The DNN is trained with supervised learning. The true state of the cell is identified by an expert and it is used as a label during training. The model is trained with maximum of 100 epochs and the best state of the model is used.

TABLE II. SUMMARY OF THE USED HYPERPARAMETERS AND THE BEST IDENTIFIED VALUES

| Model part | Hyperparameter | Possible values | Best value |
|---|---|---|---|
| Clustering block | Clustering algorithm | K-Means and Gaussian Mixture Models | - |
| Clustering block | Cluster count | 2–10 | - |
| DNN | Cell max samples | 25, 50 or 100 | 100 |
| DNN | Units per layer | 25, 50 or 100 | 100 |
| DNN | Hidden layers | 0, 1 or 3 | 4 |
| DNN | Decreasing units | False or True | True |
| DNN | Optimizer | SGD | - |
| DNN | Activation | ReLu | - |
| DNN | Learning rate | 0.001 | - |
| DNN | Loss | MSE | - |
| DNN | Regularization of the last layer | None or L1-regularization | None |

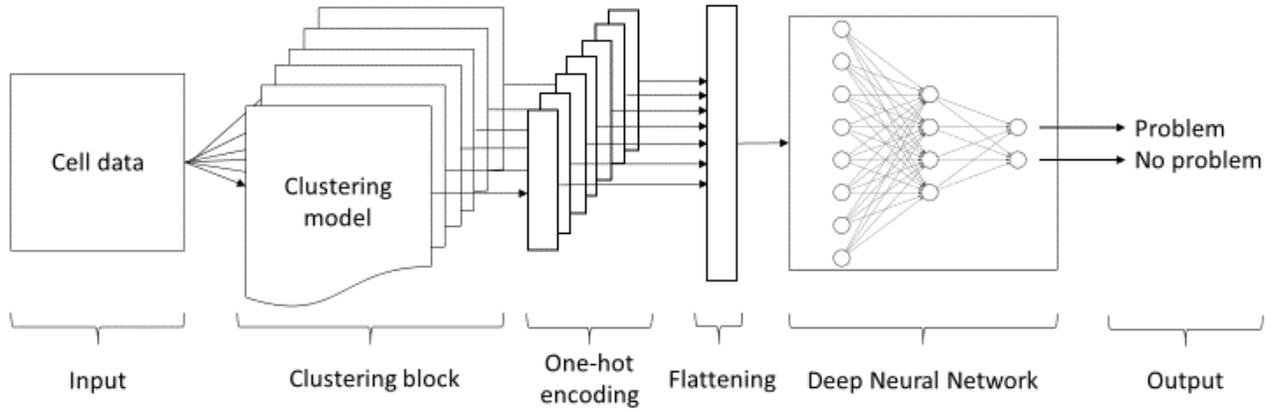

Fig. 4. Illustration of model architecture. Cell data is used as an input for the clustering block which training is shown in Fig. 2. The clustering block clusters the data in multiple different ways. The output of the clustering block is one-hot encoded, flattened and then used as an input of the DNN which is trained with cell labels provided by the expert. The DNN makes the final decision whether the cell is problematic or not. Expert labels are only required during model training, and after training the model can classify new cells on its own. The model is described in greater detail in the text.

## V. EXPERIMENTS AND RESULTS

### A. Baseline

We create a simple model for identifying problematic cells to work as a baseline for our proposed method. The baseline is created based on a hypothesis that problematic cells share certain characteristics. The characteristics are, that the cell has problems when throughput is low and CQI is high, as usually high CQI correlates with high throughput in normal situations. Based on this assumption, the following evaluation is done to classify the cells into problematic and normal ones.

1. Calculate average of CQI and average of throughput over the whole dataset.
2. If more than 50 % of samples in a cell have lower throughput than the calculated average and higher CQI than the calculated average, identify the cell as problematic, otherwise identify as normal.

An example of a cell after labelling its data points is shown in Fig. 5. The cell is identified as normal, as less than 50 % of the samples of the cell did not exceed both thresholds.

### B. Results

Hyperparameters and the best parameters of the DNN after hyperparameter tuning are shown in Table II. We used dataset which had 25 591 samples of the data transmission between UE and cells in 4G mobile network. Table III describes the problematic cells identified by expert and prior assumption. When 70 % training and 30 % test split was used, the prior assumption of the model identified 7 problematic cells in the training set, creating 126 unique clustering models. When 90 % training and 10 % test split was used, the method assumed 8 of cells on the training set were problematic.

Results are summarized in Table IV. Baseline classifier was able to classify problematic cells with 0.50 precision and 0.17 recall, giving 0.25 F1-score in 70 % training and 30 % split. The proposed method received 0.50 precision with 0.83 recall, making 0.67 F1-score. The Area Under a Curve (AUC) of the Precision-Recall Curve (PRC) was 0.785 which is shown in Fig. 6. We see the proposed method clearly exceeded the baseline.

When the training set size was increased to 90 % of the whole dataset, the baseline had 0.50 precision with 0.14 recall, making 0.22 F1-score. The proposed method achieved 0.65 precision with 0.80 recall and 0.72 F1-score correspondingly. PRC AUC score was 0.854.

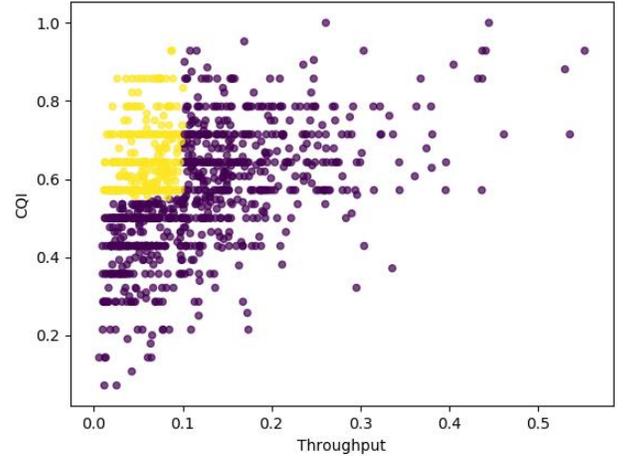

Fig. 5. Data points of a normal cell labeled by the baseline. Samples which exceeds the calculated thresholds are highlighted with yellow color. The cell is classified as normal.

TABLE III. PROBLEMATIC CELLS IDENTIFIED BY PRIOR ASSUMPTION AND THE EXPERT IN TRAINING AND TEST SETS.

|  | 70 % training set | 30 % test set | 90 % training set | 10 % test set |
|---|---|---|---|---|
| Cells | 53 | 53 | 53 | 53 |
| Problematic cells by prior assumption | 7 | - | 8 | - |
| Problematic cells by expert | 6 | 6 | 7 | 7 |

TABLE IV. PERFORMANCE RESULTS OF THE PROPOSED METHOD AND BASELINE

| Model | 70 % training and 30 % test split | | | | 90 % training and 10 % test split | | | |
|---|---|---|---|---|---|---|---|---|
| | Precision | Recall | F1-score | PRC AUC | Precision | Recall | F1-score | PRC AUC |
| The proposed method | 0.56 | 0.83 | 0.67 | 0.785 | 0.65 | 0.80 | 0.72 | 0.854 |
| Baseline | 0.50 | 0.17 | 0.25 | - | 0.50 | 0.14 | 0.22 | - |

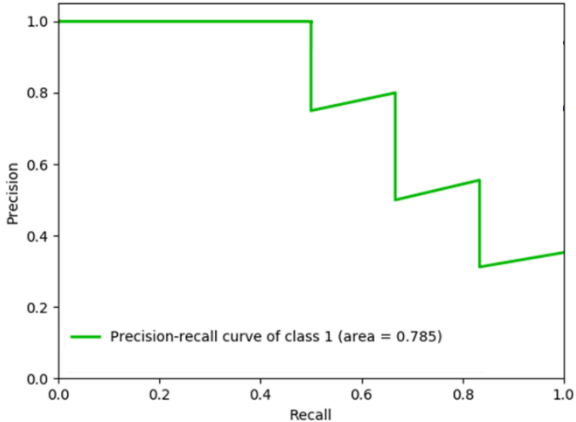

Fig. 6. The Precision Recall Curve (PRC) of the proposed method with 70 % training and 30 % test split.

Results indicate that increasing the training set size increases the performance of the proposed method. The same hyperparameters were identified as the best working ones for both tests, and these can be found from Table II. The bigger amount of test samples should be used in the latter case to make evaluation of the performance more reliable, as the average number of samples per cell was only 35. The average number of samples per cell was 157 in the 30 % test split correspondingly.

## VI. CONCLUSIONS

Identifying throughput problematic cells efficiently in mobile networks becomes an issue when the complexity of the network increases and data privacy sets limitations for the available data. To the best of our knowledge, we identified a new research area which focuses on identifying throughput problems in cells. We proposed a method which can classify cells with a low number of labels from a dataset gathered from a 4G network. The method identified problematic cells by combining clustering models and a DNN. Precision, recall and F1-score metrics were used for measuring the performance of the model. The performance of the proposed method achieved 0.67 in F1-score, exceeding 0.25 F1-score of the baseline. The results indicate that problematic cells can be identified by using only monitoring data with a low number of labels.

In this paper, we used 4G network data. However, throughput related problems are also expected to be present in 5G mobile networks, regardless of whether the network is implemented as Non-Standalone (NSA) or Standalone (SA). Future work includes using our approach with 5G network data. Furthermore, future studies could focus on measuring how a bigger training dataset affects the performance, as initial results showed a positive correlation between performance and the size of the training dataset. In addition, trying out more sophisticated clustering algorithms and a larger variety of deep neural network structures could increase performance even further.